\documentclass[11pt]{article}

\usepackage{microtype} 
\usepackage{booktabs}  
\usepackage{url}  
\usepackage{algorithm}
\usepackage{algpseudocode}
\usepackage{graphicx}

\usepackage{amsmath}
\usepackage{amsthm}

\usepackage[final,shortpaper]{automl}
\usepackage{automl}
\usepackage{cleveref}

\usepackage[%
  backend=biber,
  style=authoryear-comp,
  sortcites=true,
  natbib=true,
  giveninits=true,
  maxcitenames=2,
  maxbibnames=99,
  doi=false,
  url=true,
  isbn=false,
  dashed=false,
  uniquename=false,
  uniquelist=false,
]{biblatex}
\title{Successive Halving with Learning Curve Prediction via Latent Kronecker Gaussian Processes}

\author[1, 2]{\nameemail{Jihao Andreas Lin}{jal232@cam.ac.uk}}
\author[1]{\nameemail{Nicolas Mayoraz}{nmayoraz@google.com}}
\author[1]{\nameemail{Steffen Rendle}{srendle@google.com}}
\author[1]{\nameemail{Dima Kuzmin}{dimakuzmin@google.com}}
\author[1]{\nameemail{Emil Praun}{emilp@google.com}}
\author[1]{\nameemail{Berivan Isik}{berivan@google.com}}

\affil[1]{Google}
\affil[2]{University of Cambridge}

\hypersetup{%
  pdfauthor={}, 
  pdftitle={},
  pdfsubject={},
  pdfkeywords={}
}

\begin{document}

\maketitle

\begin{abstract}
Successive Halving is a popular algorithm for hyperparameter optimization which allocates exponentially more resources to promising candidates.
However, the algorithm typically relies on intermediate performance values to make resource allocation decisions, which can cause it to prematurely prune slow starters that would eventually become the best candidate.
We investigate whether guiding Successive Halving with learning curve predictions based on Latent Kronecker Gaussian Processes can overcome this limitation.
In a large-scale empirical study involving different neural network architectures and a click prediction dataset, we compare this predictive approach to the standard approach based on current performance values.
Our experiments show that, although the predictive approach achieves competitive performance, it is not Pareto optimal compared to investing more resources into the standard approach, because it requires fully observed learning curves as training data.
However, this downside could be mitigated by leveraging existing learning curve data.
\end{abstract}

\section{Introduction}
Hyperparameter optimization (HPO) is a critical but computationally expensive part of the machine learning workflow.
To address this problem, Successive Halving (SH) \citep{karnin2013almost,jamieson2016non} allocates computational resources efficiently across many hyperparameter candidates by operating in a series of \emph{rungs}.
In each rung, a fraction of the candidates are discarded, typically based on their current intermediate performances.
The surviving candidates are promoted to the next rung and provided with an exponentially larger resource budget.

\begin{figure}[t]
    \centering
    \includegraphics[width=6in]{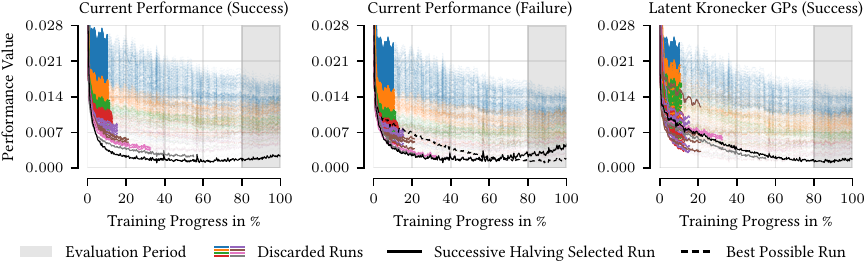}
    \caption{Visualization of Successive Halving with predictions based on current performance values (left, middle) and LKGPs \citep{lin2024scaling,lin2025scalable} (right). If intermediate performance values are indicative of future performance values, Successive Halving based on current performance values will identify the best possible run (left). However, sometimes the best possible run is discarded because its intermediate performance seems unpromising compared to other candidates at the same stage (middle). Using LKGPs to predict future performance values can identify the best possible run, despite unpromising intermediate performance values (right).}
    \label{fig:illustration}
\end{figure}

However, SH typically assumes that intermediate performance values are indicate of future performance values, which is a notable weakness.
The heuristic fails for \emph{slow starters} with seemingly unpromising intermediate performance, because SH would incorrectly discard them (see \Cref{fig:illustration}, middle).
This raises the question of whether we can improve SH by making more informed pruning decisions based on predicted performance values, instead of relying on current performance values.

In this work, we investigate SH using learning curve predictions based on Latent Kronecker Gaussian Processes (LKGPs) \citep{lin2024scaling,lin2025scalable}, a scalable probabilistic method, to predict future performance values given a set of fully and partially observed learning curves as training data.
Our hypothesis is that using these predictions to guide SH could lead to more robust and reliable identification of the best hyperparameter configurations (see \Cref{fig:illustration}).

We test our hypothesis via a large-scale empirical study using the 1TB Criteo click prediction dataset \citep{criteo_data}.
By simulating thousands of SH runs, we compare the standard approach based on current performance values to predictions based on LKGPs.
Our experiments show that, although SH with predictions based on LKGPs achieves competitive performance, the approach is not Pareto optimal compared to investing more resources into the standard SH algorithm, because it relies on additional fully observed learning curves which are used as training data.
Nonetheless, we believe that this drawback can be overcome by leveraging existing learning curve data, which opens avenues for future research.

\section{Background}
Let there be $N$ candidates, identified by their index $i = 1, ..., N$ and corresponding hyperparameters $x_i \in \mathcal{X}$.
Each candidate is a machine learning model which is trained using iterative optimization on a fixed sequence of data that is shared across all candidates.
In particular, the iterative training procedure generates performance metrics $y_{i,t}$ for each candidate $i$ and time steps $t = 1, ..., T$.

We define the best candidate and its performance as $i_* = \arg \min_i \, \mathrm{perf}(i)$ and $y_* = \min_i \, \mathrm{perf}(i)$ where $\mathrm{perf}(i) = \frac{1}{\Delta} \sum_{t=T - \Delta}^T y_{i,t}$.
In other words, the best candidate minimizes the performance metric averaged over a final window of size $\Delta$.
Our goal is to identify the best candidate $i_*$ while minimizing the number of observed performance metrics, because the latter corresponds to consumed resources.
For simplicity, we assume that all time steps are shared across all candidates and spaced uniformly, and that it takes the same amount of resources to produce any performance metric $y_{i,t}$.

\paragraph{Successive Halving}
Successive Halving (SH) was originally proposed to identify the best arm in a stochastic multi-armed bandit setting \citep{karnin2013almost}.
The algorithm was later applied to perform hyperparameter optimization in a non-stochastic setting \citep{jamieson2016non}.
The core principle of SH is to allocate fewer resources to unpromising candidates and exponentially more resources to promising ones.
In particular, the allocation schedule operates in discrete \emph{rungs}.
At each rung, $1 / \eta$ of the remaining candidates are promoted to the next rung and provided with more resources (see \Cref{alg:successive_halving}).
Typically, decisions about candidate promotions are made based on current performance values.
However, current performance values are not always indicative of future performance values given more resources.
For example, a slow starter which improves steadily may eventually overtake other candidates which converge quickly.
Therefore, the ability to accurately predict future performances could potentially improve the performance of SH.

\paragraph{Latent Kronecker Gaussian Processes}
Recently, \citet{lin2024scaling,lin2025scalable} proposed Latent Kronecker Gaussian Processes (LKGPs), a scalable probabilistic regression method which can be used to predict learning curves by considering the function $f: \mathcal{X} \times \mathcal{T} \to \mathcal{Y}$, mapping the Cartesian product space of hyperparameters $\mathcal{X}$ and time steps $\mathcal{T}$ to performance metrics.
In particular, LKGPs model $f$ using a product kernel $k((x, t), (x', t')) = k_X(x, x') k_T(t, t')$ and make nonparametric predictions based on correlations with observed training data instead of relying on specific parametric functions \citep{domhan2015speeding} or suitable synthetic data \citep{adriaensen2023}.

\begin{algorithm}[t]
\caption{Successive Halving}
\label{alg:successive_halving}
\begin{algorithmic}
\Require Initial number of candidates $N$, final number of candidates $F$, reduction rate $\eta$
\State $S \gets \lceil \log_\eta ( N / F ) \rceil$ \Comment{Calculate total number of rungs}
\For{$s = 1, \dots, S$}
    \State $R \gets (\eta^s - 1)/(\eta^S - 1) \times 100\%$ \Comment{Calculate relative resource budget for the current rung}
    \State Train remaining candidates until $R\%$ of the maximum resource budget per candidate
    \State Rank remaining candidates based on their current or predicted performance values
    \State Promote top $N / (\eta^{s - 1})$ candidates to the next rung
\EndFor
\State \Return Top $F$ candidates
\end{algorithmic}
\end{algorithm}

\section{Contributions}
In this paper, we compare the downstream performance of SH with promotion decisions based on current performance values to learning curve prediction via LKGPs.
To this end, we first create learning curve data by training neural networks, and then use the resulting data to simulate SH.

\paragraph{Creating Learning Curve Data}
We train neural networks on a fixed sequence of data, which consists of chronological click feedback data from online advertisements.
In particular, we use the Criteo 1TB dataset \citep{criteo_data} which contains a subset of Criteo's traffic over 24 days.
Each data point indicates whether or not a certain display advertisement has been clicked.

We consider Factorization Machines (FMs) \citep{rendle2010factorization}, Deep \& Cross Networks (CNs) \citep{wang2017deep, wang2021dcn}, and Mixture of Experts (MoEs) \citep{shazeer2017outrageously}.
For each architecture, we explore configurations via random search over optimization parameters such as learning rate or weight decay, and grid search over architectural parameters, such as number of hidden layers.

Since chronological data from online traffic contains strong time-dependent trends, we consider the difference to a reference model instead of raw performance values.
We construct a reference model by training a particular model for two passes over the training data and using the second pass as reference model.
The reference model is shared across all architectures and configurations.

\paragraph{Simulating Successive Halving}
To obtain statistically significant results, we simulate SH using random subsets of 256 learning curves from a total of 512 learning curves per architecture.
We set the reduction factor $\eta$ to 2 and delay SH until 10\% of the training run has finished to reduce volatile decisions.
Additionally, we set the final number of candidates to $F \in \{1, 2, 4, 8, 16, 32, 64 \}$, which leads to different performance versus compute tradeoffs.

We predict candidate rankings either based on current performance values or using LKGPs.
For the former, we take an average over the latest 20\% of the full run to improve robustness.
For LKGPs, we use squared exponential kernels with length scales per dimension, a shared kernel amplitude, and a homoscedastic noise parameter.
These parameters are optimized by maximizing the marginal likelihood using Adam with a learning rate of 0.1 and 100 iterations.
We normalize the inputs to $[0, 1]$ and standardize the outputs by subtracting the mean over the final time step and dividing by the standard deviation over all time steps.
To rank the candidates, we predict the mean $\mu_i$ and variance $\sigma^2_i$ of $\mathrm{perf}(i)$ for each candidate using 64 posterior samples.
The ranking is obtained by calculating and sorting the expected number of wins per candidate in a pairwise comparison,
\begin{equation}
    \mathbb{E}\, \mathrm{wins}(i)
    = \frac{1}{n - 1} \sum_{j \neq i} \mathbb{P}[\mathrm{perf}(i) > \mathrm{perf}(j)]
    = \frac{1}{n - 1} \sum_{j \neq i} \Phi \Big(\frac{\mu_i - \mu_j}{\sqrt{\sigma^2_i + \sigma^2_j}}\Big),
\end{equation}
where $\Phi$ is the cumulative distribution function of the standard normal distribution.
Since LKGPs uses fully observed learning curves as training data, we provide $C \in \{ 8, 16, 32, 64 \}$ curves, selected uniformly at random from the subset of 256 curves, and perform SH on the remaining curves.

\begin{figure}[t]
    \centering
    \includegraphics[width=6in]{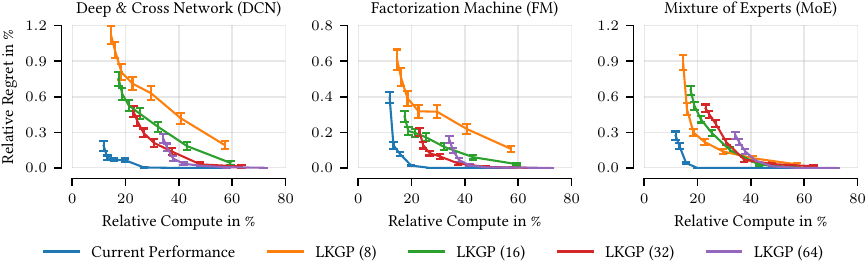}
    \caption{Comparison of Successive Halving using predictions based on current performance values (blue) versus LKGPs with varying numbers of learning curves as training data (orange, green, red, purple) for three neural network architectures (DCN, FM, MoE). As expected, increasing the amount of compute used generally improves the performance. Although LKGPs achieve competitive performance, they are not Pareto optimal due to requiring training data.}
    \label{fig:regret_vs_compute}
\end{figure}

\paragraph{Performance Evaluation}
To compare the performance of predictions based on current performance values with predictions using LKGPs, we consider the relative regret with respect to the reference model and the relative amount of compute used.
The relative regret is defined as the absolute regret divided by the performance of the reference, where the absolute regret is calculated by subtracting the performance of the best candidate proposed by SH from the best possible performance.

The relative amount of compute is calculated by dividing the number of observed performance values by the number of all performance values $N \times T$, where each performance value represents a unit of compute.
In particular, the number of observed performance values includes the initial 10\% grace period for all candidates and subsequent observations during SH.
For LKGPs, they additionally include the fully observed learning curves which are used as training data.

\paragraph{Results}
\Cref{fig:regret_vs_compute} shows the mean and standard error of the relative regret over 100 trials as a function of the relative amount of compute used.
Variations in the latter are due to different SH schedules, which are parameterized by the final number of candidates $F$.
For LKGPs, the amount of compute used is also influenced by the number of fully observed curves used as training data $C$.

As expected, using more compute by increasing $F$ generally leads to lower regret.
Interestingly, increasing $F$ seems to be particularly effective for SH with current performance values.
In this case, SH consistently achieves zero regret across all 100 trials for $F >= 32$.
The performance of SH using predictions based on LKGPs also improves as either $F$ or $C$ are increased.
However, requiring fully observed curves as training data significantly increases the amount of compute used, such that SH with LKGPs is not Pareto optimal.
Across all experiments, the same amount of compute can achieve a lower regret by instead investing compute into increasing $F$ for SH with current performances.

\section{Conclusion}
Our investigation suggests that SH with predictions based on LKGPs is not Pareto optimal compared to investing more compute into SH with predictions based on current performances.
However, this is primarily the case because LKGPs require fully observed learning curves as training data, which significantly increases the amount of compute used.
This downside could be negated by leveraging existing learning curves, which is straightforward for the same neural network architecture and motivates future work to identify ways to transfer between architectures.
Additionally, we selected the training data for LKGPs uniformly at random, which is most likely suboptimal.
Furthermore, the prediction quality of LKGPs could potentially be improved by considering specialized kernels and more elaborate feature engineering, such as input warping or embeddings.




\printbibliography

\end{document}